\documentclass{article}
\usepackage{spconf,amsmath,graphicx}
\usepackage[ruled, vlined, linesnumbered]{algorithm2e}
\pdfoutput=1
\usepackage{setspace}
\usepackage{ragged2e} 
\usepackage{booktabs,makecell, multirow, tabularx}
\usepackage{subfigure}
\usepackage{graphicx}
\usepackage{cite}
\usepackage[T1]{fontenc}

\usepackage{amsfonts}
\usepackage{algorithmic}
\usepackage{array}
\usepackage{textcomp}
\usepackage{stfloats}
\usepackage{url}
\usepackage{verbatim}
\usepackage{svg} 
\usepackage{balance}
\usepackage{diagbox}
\usepackage{slashbox}
\usepackage{caption}
\usepackage[numbers,sort&compress]{natbib}
\usepackage{color}
\usepackage{threeparttable}
\usepackage{appendix}

\newcommand{\setParDis}{\setlength {\parskip} {-0.1cm} }
\newcommand{\setParDef}{\setlength {\parskip} {0pt} }

\title{Double Reverse Regularization Network Based on Self-Knowledge Distillation for SAR Object Classification} 

\name{Bo Xu$^{*}$, Hao Zheng$^{*}$, Zhigang Hu$^{\S}$,Liu Yang, Meiguang Zheng, Xianting Feng, Wei Lin\thanks{$^{*}$Co-first authors. $^{\S}$Zhigang Hu is the corresponding author.}}

\address{School of Computer Science and Engineering,
    Central South University,
    China}

\begin{document}

\maketitle

\begin{abstract}

In current synthetic aperture radar (SAR) object classification, one of the major challenges is the severe overfitting issue due to the limited dataset (few-shot) and noisy data. Considering the advantages of knowledge distillation as a learned label smoothing regularization, this paper proposes a novel Double Reverse Regularization Network based on Self-Knowledge Distillation (DRRNet-SKD). Specifically, through exploring the effect of distillation weight on the process of distillation, we are inspired to adopt the double reverse thought to implement an effective regularization network by combining offline and online distillation in a complementary way. Then, the Adaptive Weight Assignment (AWA) module is designed to adaptively assign two reverse-changing weights based on the network performance, allowing the student network to better benefit from both teachers. The experimental results on OpenSARShip and FUSAR-Ship demonstrate that DRRNet-SKD exhibits remarkable performance improvement on classical CNNs, outperforming state-of-the-art self-knowledge distillation methods.

\end{abstract}

\begin{keywords}
Regularization, Self-Knowledge Distillation, synthetic aperture radar (SAR), SAR object classification.
\end{keywords}

\section{Introduction}
Synthetic Aperture Radar (SAR) implements all-day, all-weather observation of the earth using the synthetic aperture principle to achieve high-resolution microwave imaging. It is vital to accurately and quickly classify the ships in SAR images in performing some sea surface missions. However, SAR ship classification using data-driven deep learning faces a more significant overfitting challenge compared to optical images, mainly due to the few-shot SAR ship and noisy data.

With the rapid progress of deep learning in image processing, convolutional neural networks (CNNs) have gained increasing popularity in the field of SAR ship classification \cite{zhang2021hog,zheng2022metaboost,zhang2022polarization,zheng2023multi}. However, the complex CNN model may introduce redundant features that will further amplify the risk of overfitting. The knowledge distillation (KD) \cite{hinton2015distilling} that transfer knowledge from a cumbersome pre-trained teacher model to a lightweight student model has benn widely utilized in various SAR tasks \cite{wang2021boosting,zhang2020lossless,chen2020learning,quan2022self,zhang2023improving,xia2022dml,kang2022disoptnet,lee2021privileged}. All of the works inherit the idea of knowledge transfer from the traditional KD. Recently, \cite{yuan2020revisiting} attributed the success of KD to the regularization effect of soft labels provided by teacher model from the LSR perspective, revealing the great potential of applying KD in the field of regularization. 

\begin{figure}[tb]
	\centerline{\includegraphics[scale=0.4]{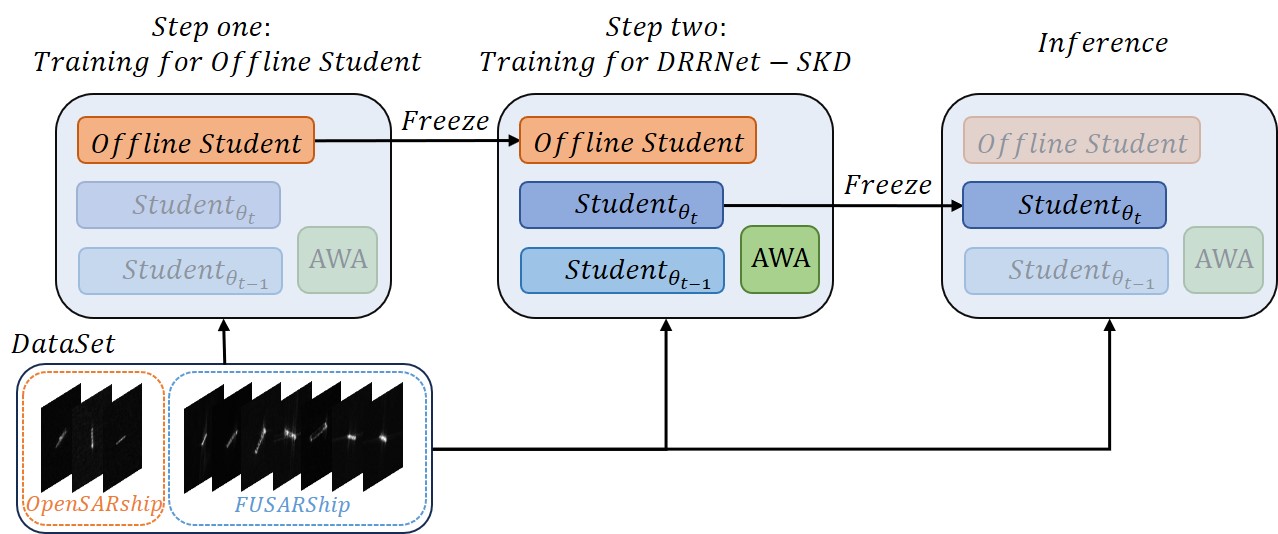}}
	\caption{The overall framework of DRRNet-SKD. The training is divided into two phases. The first stage is training the pre-trained model via DLB, whose parameters will be frozen as offline teacher in the next step. The second phase is training a new student model from scratch through DRRNet-SKD.}
	\label{DRNet.fig}
\end{figure}

To improve the generalization of SAR ship classification, this paper introduce a double reverse regularization network which incorporates both online and offline distillation. As shown in Fig. \ref{DRNet.fig}, our DRRNet-SKD employs three same-size models but different objective functions, which are the pre-trained model (Offline Student), the student model, and the last batch student model. Since the pre-trained model needs to learn the corresponding SAR ship knowledge from scratch, our framework has a two-stage training procedure. Then the well-trained student model can perform inference independently. The main contributions of this paper are as follows:
\begin{itemize}
	\setParDis
	\item [$\bullet$] 
	The effective dynamic weight assignment is found by analyzing the process of distillation and experimental validation. Concretely, increasing the weight of self-distillation and decreasing the weight of offline distillation dynamically based on the number of epoch provide more effective training for students than fixed weight.
	\item [$\bullet$]
	The Adaptive Weight Assignment (AWA) module that can adaptively assign the distillation weight based on the network performance is proposed for the first time to control the reverse change of offline and online distillation weights, which can better utilize the knowledge of both teachers.
	\item [$\bullet$]
	DRRNet-SKD achieves state-of-the-art classification accuracy compared to LSR and self-knowledge distillation approaches on the OpenSARShip and FUSAR-Ship datasets.
	\setParDef
\end{itemize}

\begin{figure*}[tb]
	
	\centerline{\includegraphics[scale=0.54]{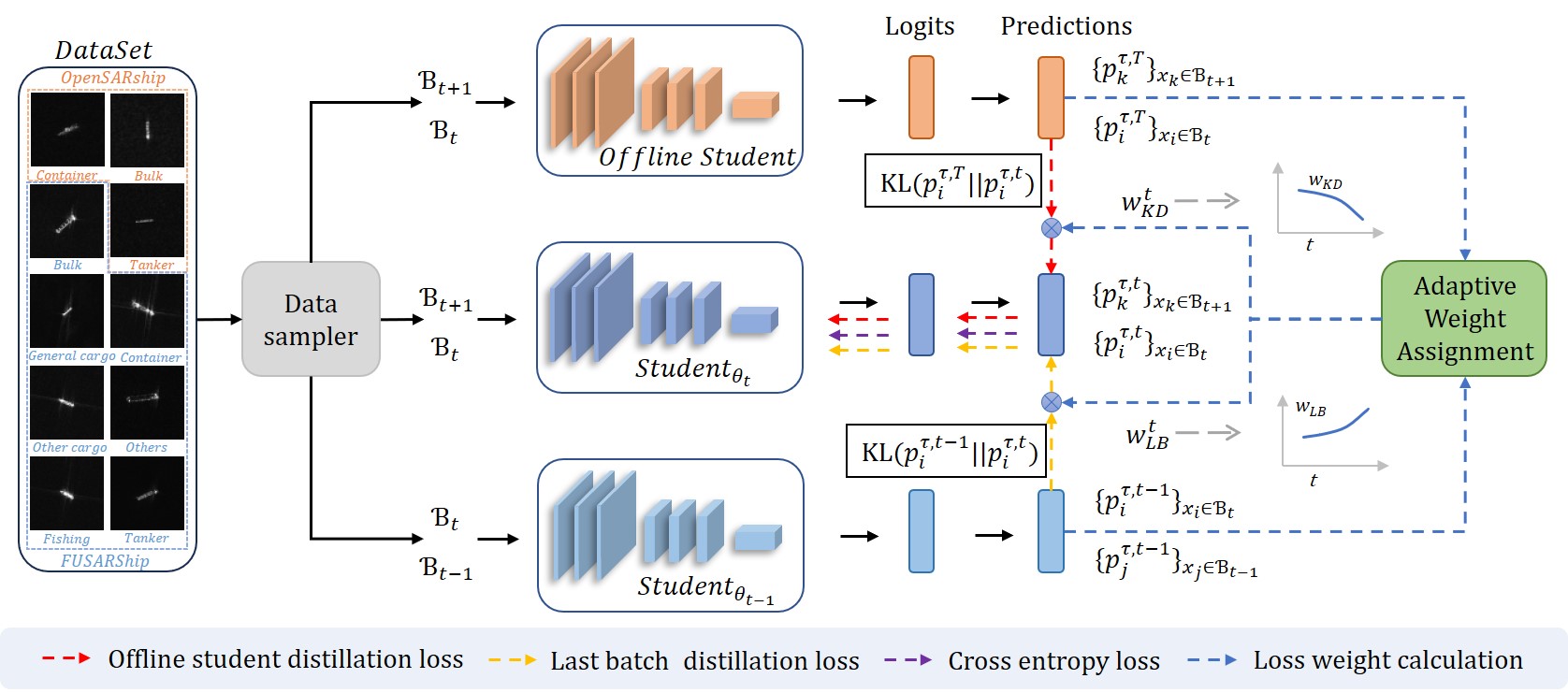}}
	\caption{The overall losses of our framework DRRNet-SKD. The $\mathcal{B}_{t}$, ${\theta_{t}}$, $p^{\tau,t}$ represent a mini-batch of data samples, trainable parameters and the soften outputs indexed in the $t^{\text {th }}$ iteration. The offline student and last batch student impose double reverse regularization on the current batch student training with the assistance of adaptive weight assignment module, respectively.}
	\label{DRNet-loss.fig}
\end{figure*}

\section{Methodology}
Fig. \ref{DRNet-loss.fig} illustrates the overall losses of DRRNet-SKD. The soft targets generated by offline student and the last batch student are exploited together to guide the student training. The trends of the two distillation weights controlled by the AWA module are depicted by the $w_{KD}$ and $w_{LB}$. The reverse variation of weight is attributed to the different variations of data knowledge contained in the soft targets supplied by both teachers. In particular, AWA module is employed to adaptively assign reverse weights of the two Kullback-Leibler (KL) divergence losses based on the soft targets rather than other factors unrelated to network performance. With this double-reverse design, DRRNet-SKD can always provide the student with superior soft targets.

\subsection{Rethinking the Weight of Knowledge Distillation}

In the offline distillation that uses the pre-trained model as its offline teacher \cite{yuan2020revisiting}, the student in the early stage of training contains limited knowledge about the data. Meanwhile, the teacher has abundant knowledge about the data and can provide high-quality soft targets for the student to learn. As student continue to learn more, the fixed soft targets provided by teacher become less beneficial to student training compared to earlier stage. Similar analysis can also be conducted in self-knowledge distillation. Thus, it can be inferred that the weight of offline distillation should decrease and the weight of self-distillation should increase as student performance grows.

\begin{table}[tb] 
	\renewcommand\arraystretch{1.0}
	\centering
	\caption{Results of fixed and unfixed distillation weights for DLB and Tf-KD on the OpenSARShip dataset. We run each method 5 times and report the “mean±std” accuracy.}
	\label{table_fixed_unfixed}
	\resizebox{\linewidth}{!}{
		\begin{tabular}{cccc}
			\hline
			\hline
			
			BackBone & Method &KD Weight & \textbf{Accuracy(\%)} \\
			\hline
			\multirow{7}{*}{ VGG-11\cite{simonyan2014very}} & Baseline & Fixed : 0 & 73.07±2.38\\ 
			
			\cline{2-4}
			
			& \multirow{3}{*}{DLB\cite{shen2022self}} & Fixed: 0.3  & 76.69±1.56\tiny{(+3.62)} \\
			& & Fixed: 0.7  & 77.82±1.13\tiny{(+4.75)} \\
			& & Unfixed:$\alpha_{t}$ & \textbf{78.14±0.37\tiny{(+5.07)}}  \\

			\cline{2-4}

			& \multirow{3}{*}{Tf-KD\cite{yuan2020revisiting}} & Fixed: 0.3  & 74.24±1.19\tiny{(+1.17)}  \\
			& & Fixed: 0.7  & 74.67±1.72\tiny{(+1.60)}   \\
			& & Unfixed:$\alpha_{t}^{\prime}$ & \textbf{75.15±0.24\tiny{(+2.08)}}  \\
			
			\hline
			\hline
		\end{tabular}
	}
\end{table}

We design the experiment to verify the inferences. The VGG-11 is trained for 100 epochs on OpenSARShip via DLB \cite{shen2022self} and Tf-KD \cite{yuan2020revisiting}, respectively. The other experimental settings are the same as those in section \ref{settings_section}. Due to the slow convergence of VGG-11, the strategy for DLB is setting a fixed weight of 0.5 for the first 50 epochs. Finally, the weights of the $t^{th}$ epoch for DLB and Tf-KD are calculated respectively according to the following formula:
\begin{equation}\label{Eq:epoch}
	\setlength{\abovedisplayskip}{3pt}
	\setlength{\belowdisplayskip}{3pt}
	\alpha_{t}=\frac{t}{T},  \alpha_{t}^{\prime}=1 - \frac{t}{T},
\end{equation}
where $T$ is the total number of epoch.

The results are shown as Table \ref{table_fixed_unfixed}. The best improvement of 5.07\% and 2.08\% is achieved by DLB and Tf-KD when the fixed weight is replaced by $\alpha_{t}$ and $\alpha_{t}^{\prime}$, respectively. Moreover, the significantly reduced accuracy variance indicates that the dynamic weight can enhance the stability and generalization of classification model. But the adoption of dynamic weight also brings some new challenges. Each of them presents a high risk of overfitting when the distillation weight is small. And the weight update strategy based on the number of epoch may not be reasonable enough due to the network performance not always improving with the increase of epoch.

\subsection{The Adaptive Weight Assignment (AWA) Module}

Inspired by the multi-teacher offline distillation weight assignment \cite{zhang2022confidence}, we exploit the cross-entropy between the softened teacher predictions and true labels to assign adaptive weight as follows:
\begin{equation}\label{Eq:L_ON}
	\setlength{\abovedisplayskip}{3pt}
	\setlength{\belowdisplayskip}{3pt}
	\mathcal{L}_{ON} = -\sum_{c=1}^{C} y^{c} \log\left( \varphi \left(  \boldsymbol{p}_{\theta_{t-1}} / \alpha_{\tau} \right)\right),
\end{equation}
\begin{equation}\label{Eq:L_OF}
	\setlength{\abovedisplayskip}{3pt}
	\setlength{\belowdisplayskip}{3pt}
	\mathcal{L}_{OF} = -\sum_{c=1}^{C} y^{c} \log \left( \varphi \left(\boldsymbol{p}^{T} / \alpha _{\tau} \right)\right),
\end{equation}
where $\boldsymbol{p}_{\theta_{t-1}}$ and $\boldsymbol{p}^{T}$ are the predictive distributions of ${t-1}^{th}$ iteration student and offline student, respectively. The temperature-like parameter $\alpha_{\tau}$ softens the distribution to smooth the values of the two equations. As the student continues to learn more, the value of Eq. \ref{Eq:L_ON} will be close to the value of Eq. \ref{Eq:L_OF}. So, the online and offline distillation weights at the $t^{th}$ iteration can be calculated by:
\begin{equation} \label{Eq:f}
	\setlength{\abovedisplayskip}{3pt}
	\setlength{\belowdisplayskip}{3pt}
	w^{t}_{LB} = exp( \mathcal{L}_{OF} - \mathcal{L}_{ON}),
\end{equation}
\begin{equation} \label{Eq:g}
	w^{t}_{KD} = \alpha -  w^{t}_{LB},  \alpha \geq w^{t}_{LB}.
\end{equation}
The value of Eq. \ref{Eq:f} will gradually increase and finally fluctuate around a certain number (near 1). Conversely, the value of Eq. \ref{Eq:g} will gradually decrease but never be less than zero. $\alpha$ controls the distillation weight range of offline student to represent its importance during the distillation process. So the AWA module can achieve adaptive assignment of distillation weight based on network performance.

\subsection{The Overall Loss Function of DRRNet-SKD}

In this paper, we focus on single-objective fully-supervised classification tasks.
We denote $\boldsymbol{z}_{\theta } = \left[z^{1}, \ldots, z^{C}\right]$ as the logits outputted by the network $\theta$, where $C$ is the class number.
The predicted probability of $c$-th class is calculated by a softmax function:
\begin{equation}
	\setlength{\abovedisplayskip}{3pt}
	\setlength{\belowdisplayskip}{3pt}
	p_{i} \left(c\right) =  \varphi \left( z^{c} \right)= \frac{\exp \left(z^{c} \right)}{\sum_{j=1}^{C} \exp \left(z^{j} \right)}, c \in \{1,2, \cdots, C\}.
\end{equation}
The student is trained by optimizing the Kullback-Leibler (KL) divergence loss between the softened outputs from the last mini-batch and the current mini-batch:
\begin{equation}\label{Eq:L_LB}
	\mathcal{L}_{LB} = \tau^{2} \cdot KL\left(\varphi  \left( \boldsymbol{z}_{\theta_{t-1} }   / \tau \right) \| \varphi  \left( \boldsymbol{z}_{\theta_{t}} / \tau \right) \right),
\end{equation}
where $\boldsymbol{z}_{\theta_{t-1}}$ is the  output logits of the $t-1^{th}$ iteration student. Then they could be expressed as teacher and student, respectively. $\tau$ represents the distillation temperature, which regulates the degree of smoothing of the network outputs. Similarly, the KL divergence loss between the softened outputs from the offline student and the current mini-batch is
\begin{equation} \label{Eq:L_KD}
	\setlength{\abovedisplayskip}{3pt}
	\setlength{\belowdisplayskip}{3pt}
	\mathcal{L}_{KD} = \tau^{2} \cdot KL\left(\varphi  \left( \boldsymbol{z}^{T} / \tau \right) \| \varphi  \left( \boldsymbol{z}_{\theta_{t}}/ \tau \right) \right),
\end{equation}
where $\boldsymbol{z}^{T}$ is the output logits of offline student. To train a multi-class classification model, we typically adopt the Cross-Entropy (CE) loss between the predicted and ground-truth label distributions as the loss function:
\begin{equation}
	\setlength{\abovedisplayskip}{3pt}
	\setlength{\belowdisplayskip}{3pt}
	\mathcal{L}_{C E}=-\sum_{c=1}^{C} y^{c} \log \left( \varphi\left(z_{\theta}^{c}\right)\right).
\end{equation}
Eventually, the overall loss function of our DRRNet-SKD is summarized as:
\begin{equation}
	\setlength{\abovedisplayskip}{3pt}
	\setlength{\belowdisplayskip}{3pt}
	\mathcal{L}= \mathcal{L}_{CE} + w_{LB} \cdot\mathcal{L}_{LB} +  w_{KD} \cdot  \mathcal{L}_{KD}.
\end{equation}

\section{Experiment And Result Analysis}\label{settings_section}

The experiments are performed on a server equipped with an Intel Pentium processor, an NVIDIA RTX3090 GPU and 64GB memory. The proposed DRRNet-SKD and the compared methods were all implemented using PyTorch 1.12.1 in a Python 3.8.15 environment. All the CNNs are built using the open-source PyTorch.

\subsection{Experimental Settings}

\begin{table*}[tb]
	\renewcommand\arraystretch{1.08}
	\centering
	\caption{Comparison of DRRNet-SKD (Ours) with LSR and state-of-the-art self-knowledge distillation methods. The underlined values represent the next largest accuracy gains over the Baseline, and the bold black numbers indicate the highest accuracy achieved by different backbones. We run every backbone five times and report the “mean±std” accuracy.}
	\begin{tabular}{ccccccc}
		\hline
		\hline
		
		Dataset & Method & AlexNet & VGG-11 & VGG-16 & ResNet-18 & DenseNet-121 \\
		\hline
		
		\multirow{5}{*}{{OpenSARShip}}
		& Baseline                          & 68.88±2.05                             & 73.07±2.38                           & 72.34±2.51                                & 72.15±1.25                             & 75.69±1.60  \\
		& LSR \cite{szegedy2016rethinking}  & 70.19±2.79\underline{\tiny{(+1.31)}}  & 76.94±0.86\tiny{(+3.87)}            & 74.34±1.61\tiny{(+2.00)}                 & 73.20±1.29\tiny{(+1.05)}              & 78.03±0.72\underline{\tiny{(+2.34)}}  \\
		& Tf-KD \cite{yuan2020revisiting}   & 69.35±1.22\tiny{(+0.47)}              & 74.67±1.72\tiny{(+1.60)}             & 75.04±1.73\underline{\tiny{(+2.70)}}    & 72.75±0.97\tiny{(+0.60)}              & 75.15±1.99\tiny{(-0.54)} \\
		& DLB \cite{shen2022self}           & 70.15±1.47\tiny{(+1.27)}              & 77.82±1.13\underline{\tiny{(+4.75)}} & 74.63±0.63\tiny{(+2.29)}                & 74.46±1.08\underline{\tiny{(+2.31)}}  & 76.62±1.26\tiny{(+0.93)}  \\
		& \textbf{Ours}& \textbf{70.67±0.49\tiny{(+1.79)}}     & \textbf{80.03±0.87\tiny{(+6.96)}}   & \textbf{78.15±1.72\tiny{(+5.81)}}   & \textbf{75.10±0.91\tiny{(+2.95)}}     & \textbf{78.48±0.72\tiny{(+2.79)}} \\

		\hline
		& Method  & AlexNet  & VGG-16  & VGG-19 & ResNet-18  & ResNet-50 \\
		\hline
		\multirow{5}{*}{{FUSAR-Ship}}
		& Baseline                           &   79.36±0.34   & 83.51±0.77 & 82.51±0.70 &  79.48±0.39  & 80.29±0.46\\
		& LSR \cite{szegedy2016rethinking}   &   80.31±0.36\underline{\tiny{(+0.95)}}  & 85.41±0.70\underline{\tiny{(+1.90)}} & 84.34±0.40\tiny{(+1.83)}             & 80.67±0.30\tiny{(+1.19)}             & 81.40±0.46\tiny{(+1.11)} \\
		& Tf-KD \cite{yuan2020revisiting}    &   79.60±0.76\tiny{(+0.24)}              & 84.35±0.80\tiny{(+0.84)}             & 83.24±0.88\tiny{(+0.73)}             & 80.88±0.47\underline{\tiny{(+1.40)}} & 81.91±0.35\underline{\tiny{(+1.62)}} \\
		& DLB \cite{shen2022self}            &   80.24±0.58\tiny{(+0.88)}              & 84.53±0.36\tiny{(+1.02)}             & 84.79±0.35\underline{\tiny{(+2.28)}} & 80.71±0.41\tiny{(+1.23)}             & 81.17±0.41\tiny{(+0.88)} \\
		& \textbf{Ours}&  \textbf{81.02±0.49\tiny{(+1.66)}}  & \textbf{86.07±0.54\tiny{(+2.56)}} &   \textbf{86.97±0.40\tiny{(+4.46)}}   & \textbf{81.58±0.44\tiny{(+2.10)}}  & \textbf{82.24±0.33\tiny{(+1.95)}}  \\
		
		\hline
		\hline
	\end{tabular}
	\label{tab:ab_Fusion}	
	
\end{table*}

\textbf{Data Description.} The OpenSARShip Dataset has been widely used to evaluate SAR ship classification methods since 2018. Three primary ship categories are utilized for training and testing. The FUSAR-Ship dataset has a higher resolution and a larger number of samples compared to OpenSARShip. The seven types of ships used in this experiment, i.e., fishing, bulk, container, tanker, other cargo, general cargo, and others.
\textbf{Experiment settings.} All models are trained for 100 epochs using Adam optimizer with a learning rate of 0.0002, and the learning rate is decayed by 20\% every seven epochs. The batchsize is set to 16 on OpenSARShip and 32 on FUSAR-Ship. Since Batch Normalization(BN) \cite{ioffe2015batch} has been widely used in VGG, Pytorch provides a version with BN layers that will be employed in our experiments.

\subsection{SAR Ship Classification Results}

As shown in Table \ref{tab:ab_Fusion}, the results show that our proposed method can significantly improve the performance of all the backbone networks, even on the larger and higher-resolution SAR ship dataset. It can be found that our framework is more effective in the VGG series. VGG-11 obtains the highest improvement among the backbones on OpenSARShip, achieving an accuracy gain of 6.96\% over the baseline. VGG-19 achieves the most improvement on FUSAR-Ship, increasing the accuracy by 4.46\% over the baseline. The results sufficiently demonstrate the effectiveness of our method.

\begin{figure*}[htbp!]  
	\flushleft
	
	\subfigure[ResNet-18]{
		\includegraphics[width=4.66cm]{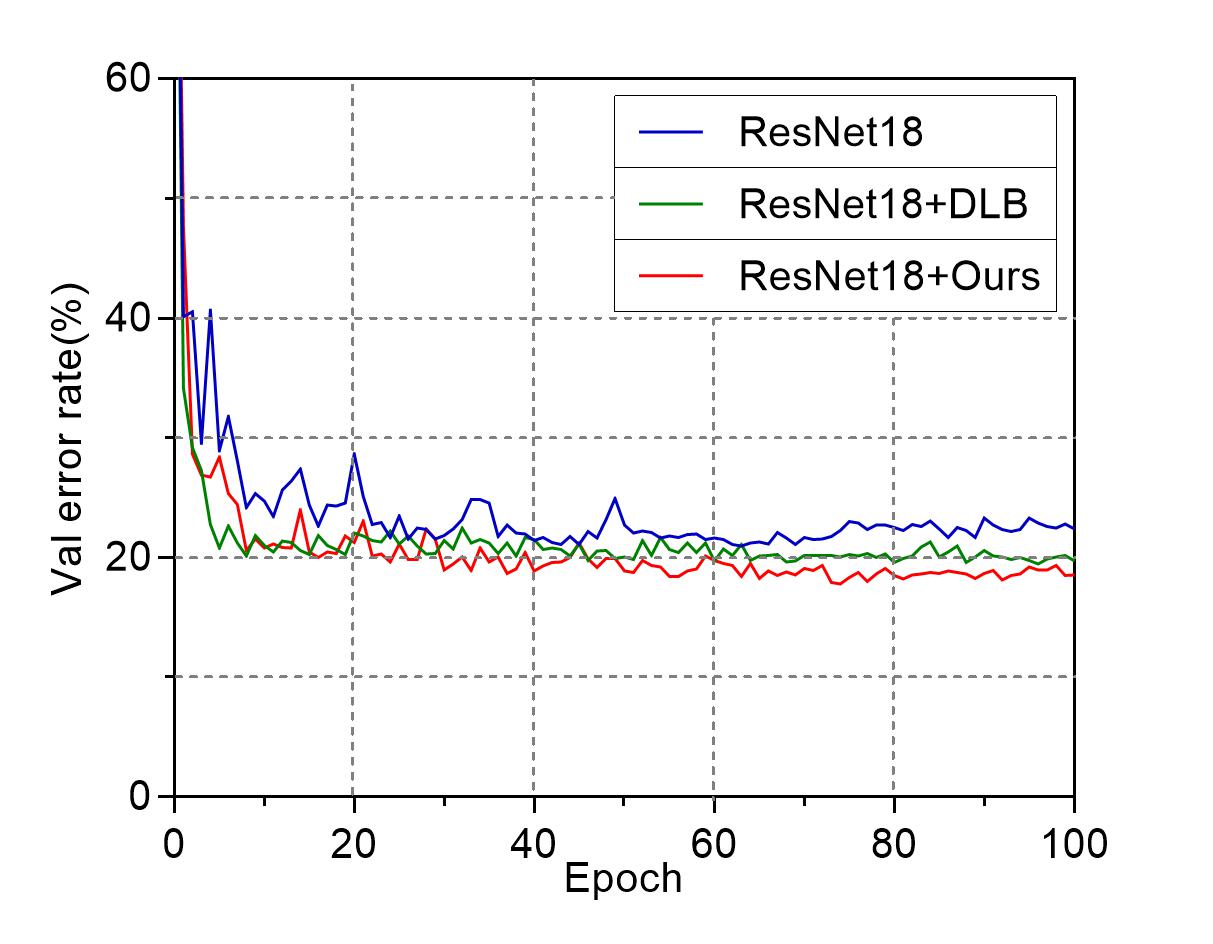}
	}\hspace{-0.6cm}
	\subfigure[ResNet-50]{
		\includegraphics[width=4.66cm]{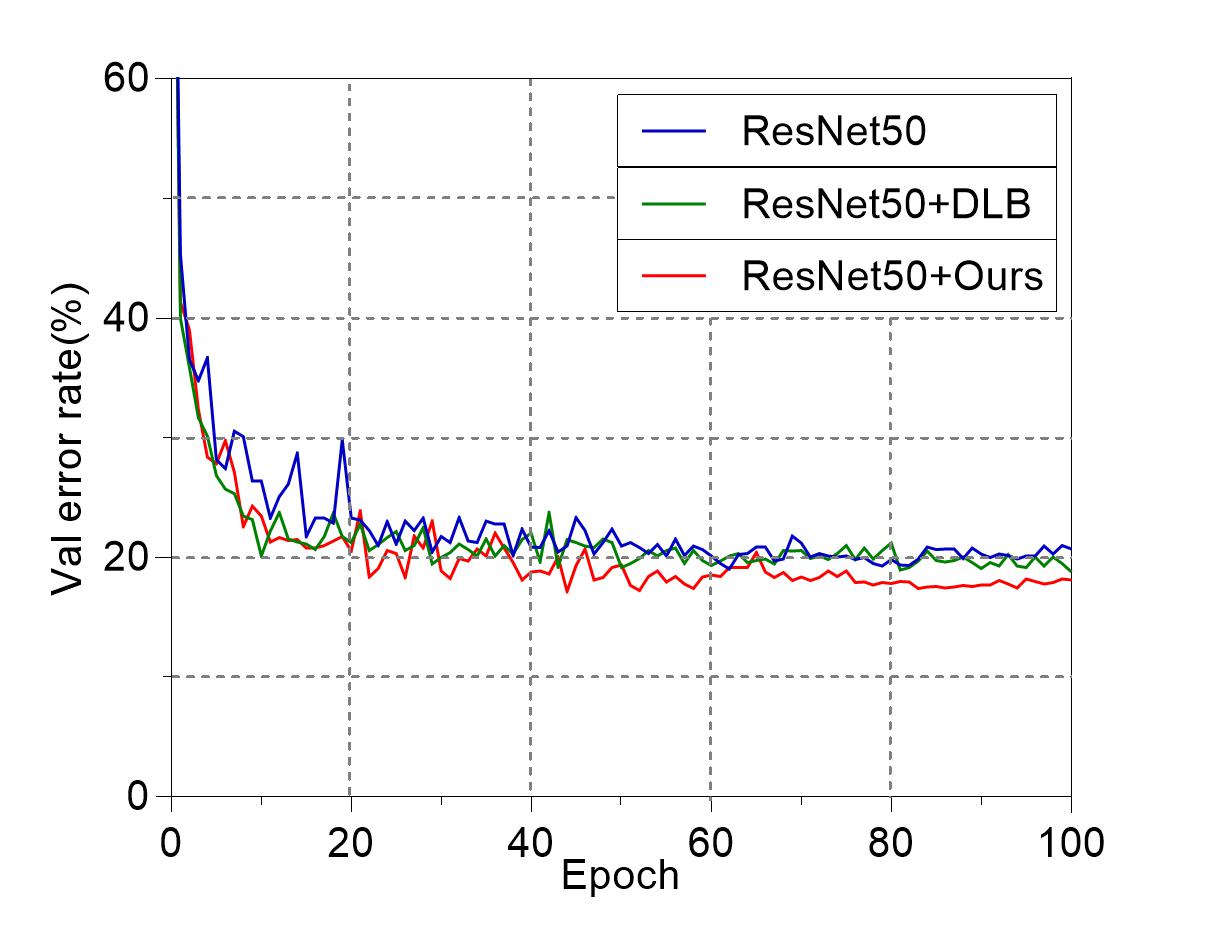}
	}\hspace{-0.6cm}
	\subfigure[VGG-16]{
		\includegraphics[width=4.66cm]{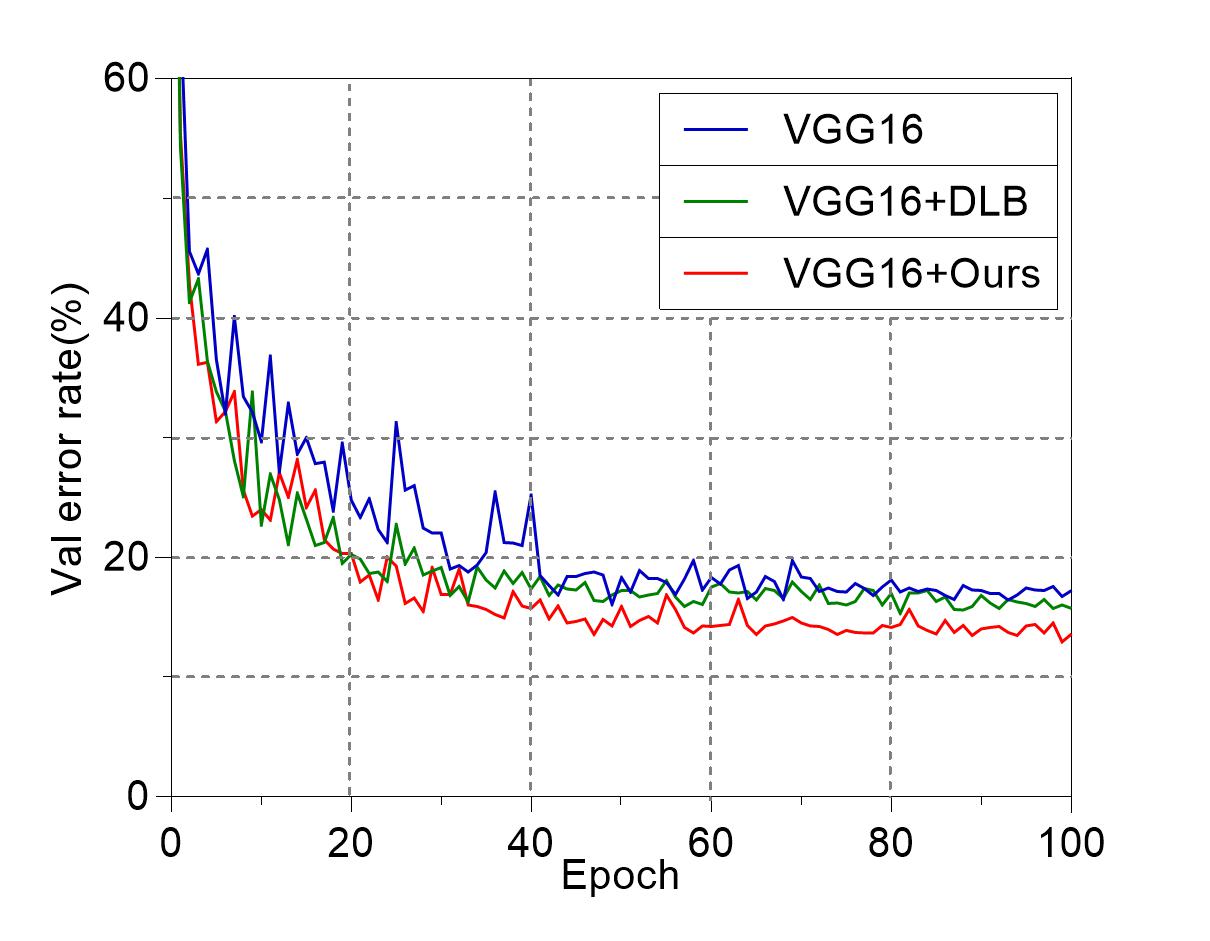}
	}\hspace{-0.6cm}
	\subfigure[VGG-19]{
		\includegraphics[width=4.66cm]{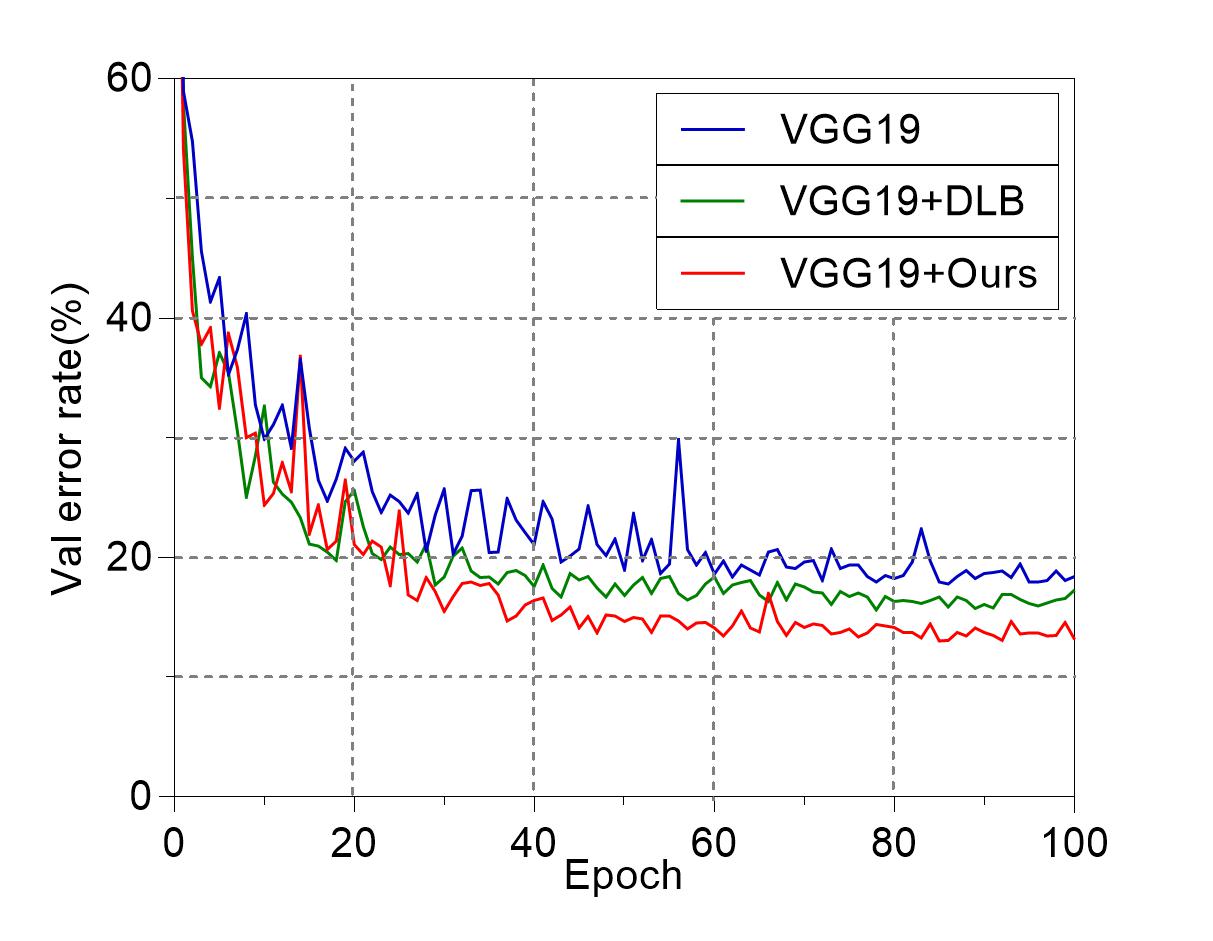}
	}
	\caption{The training performance of ResNet-18, ResNet-50, VGG-16, VGG-19 for baseine, DLB and ours on FUSAR-Ship.}
	\label{Fig:training performance}	
	\vspace{-0.2cm} 
\end{figure*}

\subsection{Comparison With the State-of-the-Art Methods}

To demonstrate the superiority of our framework, DRRNet-SKD is compared against classical LSR and the best regularization methods based on self-knowledge distillation. As shown in Table \ref{tab:ab_Fusion}, we observe that the first three regularization methods, LSR, Tf-KD, and DLB, can improve the performance of almost all backbones. However, the same regularization method does not always work best for different backbones. For instance, on FUSAR-Ship, LSR improves the accuracy of AlexNet and VGG-16 by 0.95\% and 1.90\%, respectively, surpassing the gains achieved by Tf-KD and DLB. But this improvement is not observed on the other three networks. Our DRRNet-SKD consistently can achieve the best results for different networks on both datasets, demonstrating its superior effectiveness and universality compared to other regularization methods. To provide a clearer comparison, we plotted the training performance of the four networks for baseline, DLB, and our method on FUSAR-Ship, as shown in Fig. \ref{Fig:training performance}. Compared to DLB, our method can further accelerate the convergence speed and improve the accuracy.

\section{Conclusion and Limitation}

In this research, we propose a novel double reverse regularization network based on self-knowledge distillation to improve the generalization of SAR ship classification. Firstly, our analysis and experiments show that the distillation weight that vary with the network performance is more advantageous for student training than fixed weight in some distillation methods. Secondly, we designed the Adaptive Weight Assignment (AWA) module to combine online and offline distillation in a complementary way to guide the student training. The AWA module can adaptively assign distillation weight at the sample level based on the current network performance to better utilize the knowledge from both teachers, leading to improved performance and generalization. Extensive experimental results show that our DRRNet-SKD outperforms other self-knowledge distillation approaches and exhibits optimal performance in the SAR ship classification.

\bibliographystyle{IEEEbib}
\bibliography{references}

\newpage
\begin{appendices}
\appendixpage

\setcounter{table}{0}
\setcounter{figure}{0}
\setcounter{equation}{0}
\renewcommand{\thetable}{R\arabic{table}}
\renewcommand{\thefigure}{R\arabic{figure}}
\renewcommand{\theequation}{R\arabic{equation}}

\section{Extension Experiments}

\subsection{Comparison With State-of-the-Art SAR Ship Classification Methods}

The first item of Table \ref{tab:compare} lists some other classical CNNs besides Table \ref{tab:ab_Fusion}. Among the various CNNs, DenseNet-121 achieves the highest accuracy of 75.69\% and 83.86\% on the two datasets, respectively, and its performance is still much lower than our proposed DRRNet-SKD. This result demonstrates the huge gains of our double reverse regularization network, showing the great potential of CNNs. The second item of Table \ref{tab:compare} presents the results of state-of-the-art SAR ship classification methods, the highest accuracy achieved by HOG-ShipCLSNet \cite{zhang2021hog} and DUW-Cat-FN \cite{zhang2021injection} is 78.15\% and 86.86\%, respectively, which is lower than our proposed DRRNet-SKD with the accuracy of 80.03\% and 86.97\%. Our simple and effective double reverse regularization network can outperform the CNNs fused with handcrafted features, revealing that CNN still has a great untapped potentiality even without the fusion of handcrafted features.

\begin{table}[hb]
	\renewcommand\arraystretch{1.7}
	\centering
	\caption{Experimental results of state-of-the-art SAR ship classification methods on the two datasets, respectively.}
	\label{tab:compare}
	\resizebox{\linewidth}{!}{
		\begin{threeparttable} 
			\begin{tabular}{cccc}
				\hline
				\hline
				Type & Method & OpenSARShip & FUSAR-Ship \\
				\hline
				
				\multirow{5}{*}{{CNN-only}}
				&Xception      & 73.74±0.86          & 77.29±0.38  \\
				&Inception-v4  & 72.74±0.70          & 80.50±0.37   \\
				&MobileNet-v1  & 69.91±1.08          & 77.61±0.54  \\
				&DenseNet-121  & \textbf{75.69±1.60} & \textbf{83.86±0.57} \\
				\hline
				
				\multirow{4}{*}{{Feature Fusion}} 		
				& DUW-Cat-FN \cite{zhang2021injection}    & 78                   & \textbf{86.86}  \\
				& HOG-ShipCLSNet \cite{zhang2021hog}      & \textbf{78.15±0.57}  & 86.69±0.47  \\
				& Internal FC layer \cite{zheng2023multi}    & 74.75±1.21           & 84.25±0.42  \\
				& Terminal FC layer	\cite{zheng2023multi}	  & 74.10±1.42           & 83.17±0.51  \\
	
				\hline
				\multirow{1}{*}{{Ours}}				
				& \textbf{DRRNet-SKD} 	& \textbf{80.03±0.87} & \textbf{86.97±0.40} \\
				\hline
				\hline
			\end{tabular}
			\begin{tablenotes} 
				\item The standard deviation of DUW-Cat-FN are not given in the source. 
			\end{tablenotes} 
			
		\end{threeparttable} 
	}
\end{table}

\subsection{Details of Distillation Weight Assignment}

\begin{figure}[tb]
	\centering
	\subfigure[ResNet-50]{
		\includegraphics[width=7cm]{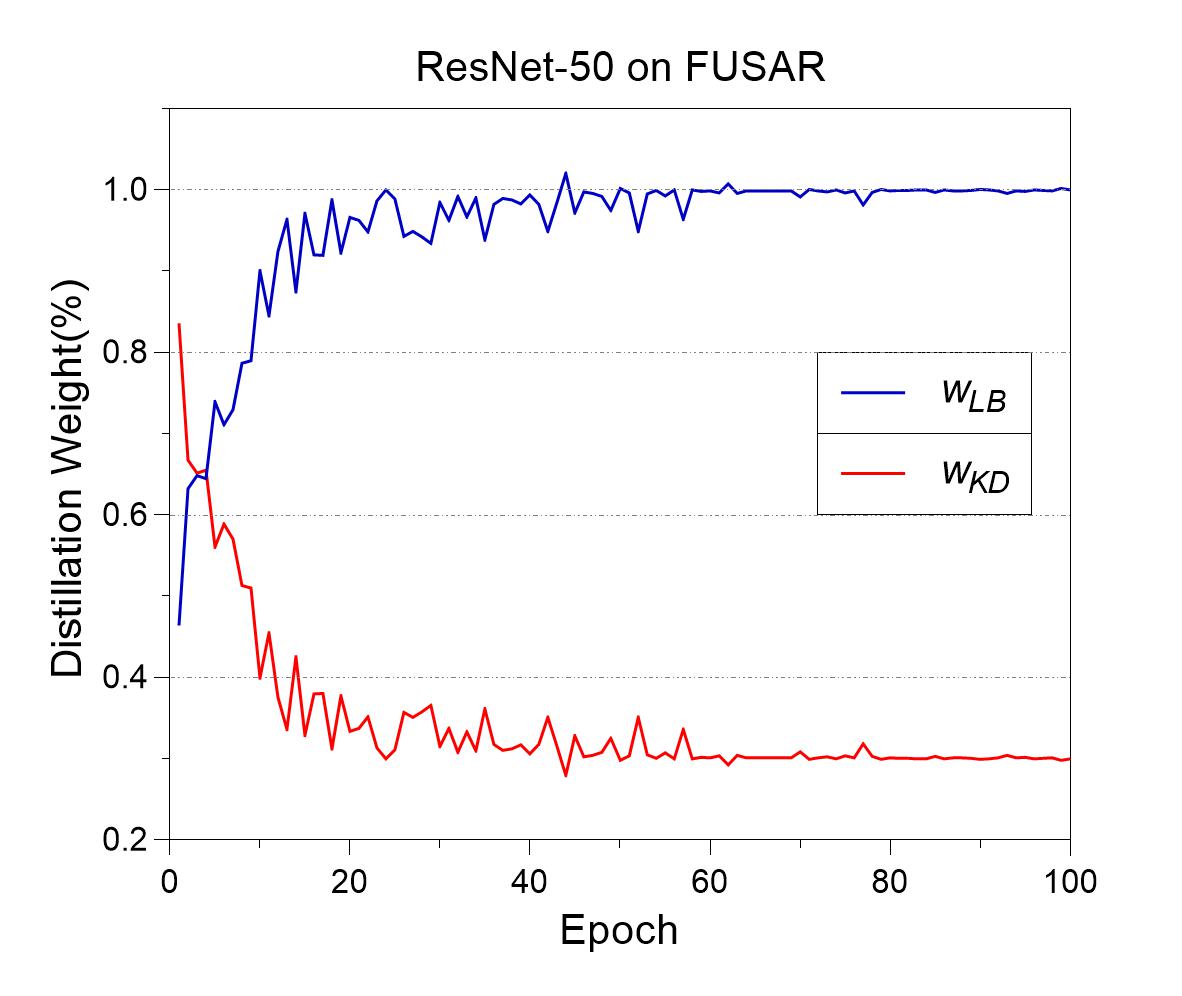}
	}
	\subfigure[VGG-19]{
		\includegraphics[width=7cm]{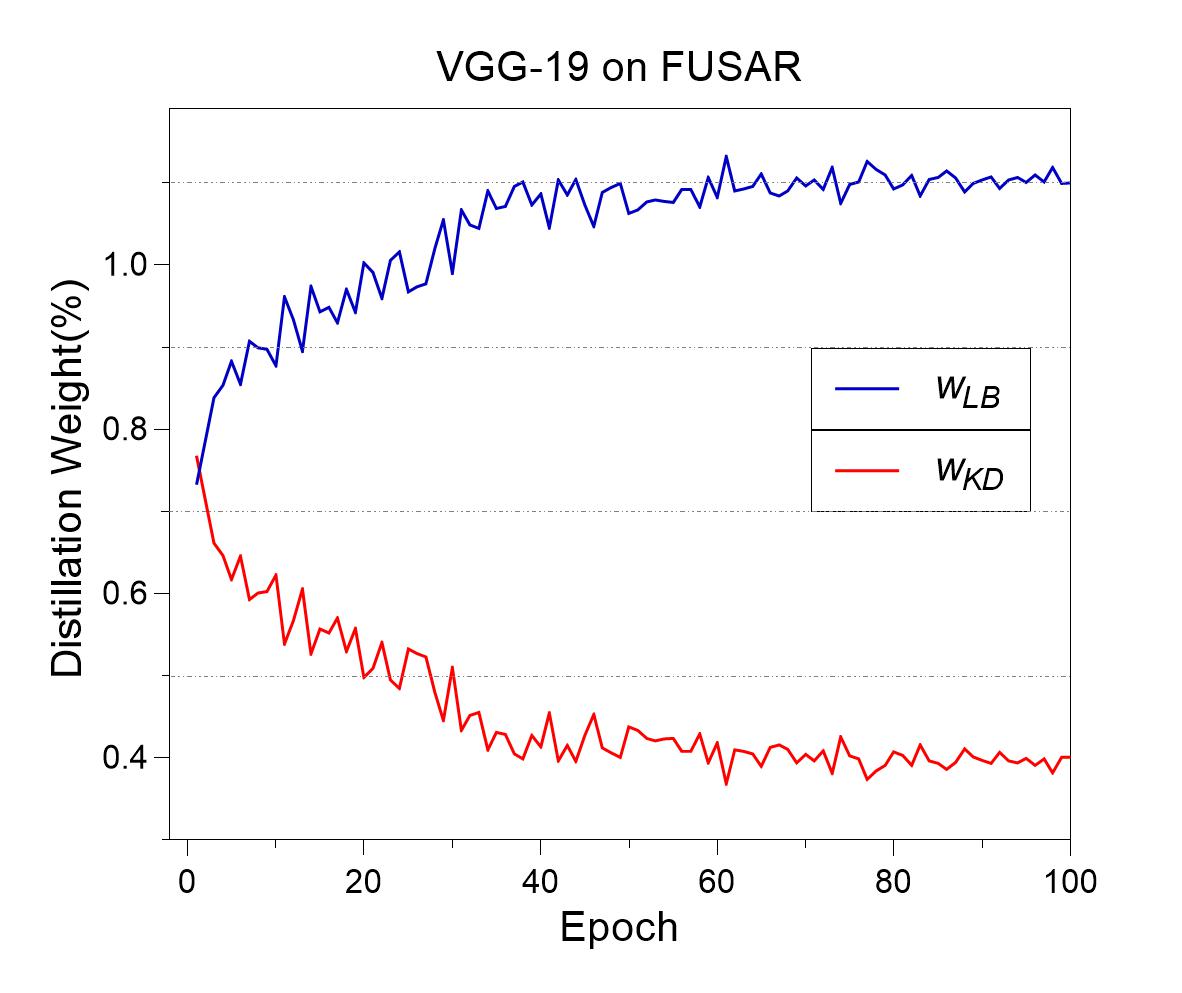}
	}
	\caption{Details of distillation weight assignment in our DRRNet-SKD on FUSAR-Ship. The alpha of AWA was set to 1.3 and 1.5 for ResNet-50 and VGG-19, respectively.}
	\label{Fig:weight}
\end{figure}

To show more details about the weight update strategy, the weights of the first batch in each epoch are sampled and plotted as shown in Fig. \ref{Fig:weight} (large sampling intervals may lead to severe fluctuation). It can be seen that the weight variation trends between ResNet-50 and VGG-19 are distinct. For ResNet-50, the two distillation weights increase or decrease rapidly during the initial 20 epochs, respectively, and then start to stabilize after the $60^{th}$ epoch. And the weights of VGG-19 change slowly and exhibit slight fluctuations throughout the training. Combined with Fig. \ref{Fig:training performance}, we see some similarities between the variation of distillation weight and its training performance. For example, the performance of ResNet-50 improves rapidly in the first 20 epochs, and the weight also changes quickly, while both the performance and the distillation weight show very stable after the $60^{th}$ epoch.

\end{appendices}

\end{document}